\documentclass[conference]{IEEEtran}
\IEEEoverridecommandlockouts
\usepackage{cite}
\usepackage{amsmath,amssymb,amsfonts}
\usepackage{algorithmic}
\usepackage{graphics}
\usepackage{lipsum}
\usepackage{grffile}
\usepackage{adjustbox}
 \usepackage{textcomp}
 \usepackage{footmisc}
\usepackage{pdfpages}
\usepackage{tabularx}
\usepackage{ifpdf}
\usepackage{epstopdf}
\usepackage{footnote}
\usepackage{url}
\usepackage{blindtext}
\usepackage{xcolor}
\usepackage{colortbl}%
  \newcommand{\myrowcolour}{\rowcolor[gray]{0.925}}
\usepackage[utf8]{inputenc}
\usepackage{babel,textcomp}
\usepackage{subfigure}
\usepackage{csquotes}

\def\BibTeX{{\rm B\kern-.05em{\sc i\kern-.025em b}\kern-.08em
    T\kern-.1667em\lower.7ex\hbox{E}\kern-.125emX}}

\newcounter{RZNumberOfComments}
\stepcounter{RZNumberOfComments}

\begin{document}

\title{ Hate Speech and Offensive Language Detection using an Emotion-aware Shared Encoder}

\author{\IEEEauthorblockN{Khouloud Mnassri, Praboda Rajapaksha, Reza Farahbakhsh, Noel Crespi}
\IEEEauthorblockA{\textit{Samovar, Telecom SudParis, Institut Polytechnique de Paris,}
91120 Palaiseau, France\\
\{khouloud.mnassri, praboda\_rajapaksha, reza.farahbakhsh, noel.crespi\}@telecom-sudparis.eu}}

\maketitle

\begin{abstract}
    The rise of emergence of social media platforms has fundamentally altered how people communicate, and among the results of these developments is an increase in online use of abusive content. Therefore, automatically detecting this content is essential for banning inappropriate information, and reducing toxicity and violence on social media platforms. The existing works on hate speech and offensive language detection produce promising results based on pre-trained transformer models, however, they considered only the analysis of abusive content features generated through annotated datasets. This paper addresses a multi-task joint learning approach which combines external emotional features extracted from another corpora in dealing with the imbalanced and scarcity of labeled datasets. Our analysis are using two well-known Transformer-based models, BERT and mBERT, where the later is used to address abusive content detection in multi-lingual scenarios. Our model jointly learns abusive content detection with emotional features by sharing representations through transformers' shared encoder. This approach increases data efficiency, reduce overfitting via shared representations, and ensure fast learning by leveraging auxiliary information. Our findings demonstrate that emotional knowledge helps to more reliably identify hate speech and offensive language across datasets. Our hate speech detection Multi-task model exhibited 3\% performance improvement over baseline models, but the performance of multi-task models were not significant for offensive language detection task. More interestingly, in both tasks, multi-task models exhibits less false positive errors compared to single task scenario.

\end{abstract}
\begin{IEEEkeywords}
	Social media, Natural Language
Processing, Hate speech, offensive
language, Twitter, BERT, Multilingual BERT, Multi-task learning, emotional knowledge, shared encoder.
\end{IEEEkeywords}

\section{Introduction}
\vspace{-5pt}
People have become addicted to social media platforms in recent decades as means of engaging and connecting with each other. Through social media platforms like Twitter, individuals are increasingly communicating and expressing their opinions, and emotions. However, their content can contain harmful information prejudiced against a certain person or group, manifested as abusive language. It is challenging to have a concrete final definition of hate and offensive language, but in general, according to the United Nations\footnote{\url{https://www.un.org/en/hate-speech}\label{this}}: “hate speech” refers to \textbf{offensive} discourse targeting group or an individual based on inherent characteristics (race, religion or gender) and that may threaten social peace.” As for offensive language, it is more general referring to any content that can implicitly or explicitly offend the other or make him uncomfortable. 
Today, it becomes challenging and even impossible to manually track the substance of posts due to the huge and unregulated content that is uploaded every day online. 
However, there have been many research attempts in automating the detection of abusive content on online platforms.
The majority of these attempts adopt supervised learning and deep learning methods that were trained using an annotated dataset\cite{fortuna2018survey}. Among the substantial drawbacks in these approaches include the lack of training data, 
and in data bias, as well as the ambiguity of the abusive content that can be challenging to accurately detect using traditional NLP methods\cite{davidson2017automated}. The latters start with using machine learning \cite{davidson2017automated}, then deep learning, transfer learning and ensemble learning \cite{sysreview2021}... Thus, we aim to involve training a single transformer-based model on multiple tasks simultaneously as it has been shown to improve performance of abusive language detection.
This paper proposes a multi-task joint learning approach that utilizes additional features (emotions), to improve model performances. Emotion categorization from the text is deeply aligned with scientific concepts that have been studied for a very long time in the framework of Sentiment Analysis (SA) \cite{medhat2014sentiment}. In fact, emotion classification seeks to automatically classify texts in a precise manner according to several classes such as anger, fear... \cite{demszky2020goemotions}. A psychology study \cite{patrick1901psychology} found a direct correlation between the speaker's psychological and emotional state, and his abusive speech \cite{cao2020deephate}. For instance, abusive content presents unpleasant attitudes and feelings like anger, disgust, fear, and sadness. Hence, in this paper, we studied new methods of improving the detection of hate speech and offensive language by integrating emotional knowledge as an additional related feature.
In the Multi-Task Learning (MTL) scenario, multiple tasks are learned in parallel while using a shared representation \cite{caruana1997multitask}. In comparison to learning multiple tasks individually and sequentially, this joint learning approach effectively increases the sample size while training a single model, which leads to improved performance by increasing its generalization \cite{zhang2021survey}.
Based on the recent state-of-the-art results given by implementing BERT in hate speech/offensive language detection \cite{sohn2019mc, mozafari2019bert, mnassri2022bert}, as well as mBERT (in cross-lingual setting) \cite{sohn2019mc, medhat2014sentiment}, we have chosen to build our MTL model using those two language models.
Following listed the main contributions of this paper:
\begin{itemize}
   \item Building a multi-task learning framework that enables the model by sharing representations between several related tasks and generalizing better by achieving better performance for the hate speech and offensive language detection task.
   \item Use of BERT and mBERT pre-trained models as shared encoders to create an MTL model, and analyze and compare their efficiency.
    \item Use of external related features (emotions) to improve hate speech detection task. We proved the hypothesis of the relationship between the spread of hateful and abusive content and the emotional psychological status of its writer.
    \item Model optimization within learning multiple tasks in parallel while using a shared transformer representation, which helps in avoiding the computation expenses and task-specific fine-tuning training step. This ensures making predictions at inference time faster than training two different models for every task independently.

\end{itemize}
 

Our joint learning approach uses a transformer-based shared encoder to implement a multi-task model for categorizing hate speech and offensive language. It aims to solve the issue of labeled data scarcity by sharing representations between several tasks, using an auxiliary dataset from the secondary task (emotion knowledge). The proposed multi-task model exhibits higher performance with fewer classification errors compared to single-task baseline models in both hate speech and offensive language detection. 

\section{Literature Survey}
\vspace{-5pt}

\subsection{Multi-task learning on hate/offensive speech}
The implementation of multitask learning approach in the field of NLP abusive language detection remains a new approach.
The first approaches used deep neural networks. Liu et al. \cite{liu2019fuzzy} proposed a three-level framework. It includes detecting: hate speech, its types, and its topics. They developed a fuzzy ensemble approach in the setting of multi-task learning considering each type of hate speech as task prediction head. Their ensemble gave a detection rate of 0.93. Moreover, Kapil et al. \cite{kapil2020deep} proposed a deep learning Shared-Private multi-task model to leverage the information from 5 abusive tasks. As a result, they built 4 deep neural networks,
 and training them on 5 datasets, they managed to get 26 models' combinations. Their approaches outperformed the single-task's with macro-F1 scores between 10\% and 27\%. Furthermore,
Abu Farah et al. \cite{farha2020multitask} worked on Arabic language within a joint learning approach, using sentiment analysis.
Their best model is a multitask learning framework, based on CNN-BiLSTM. It gave macro F1-score of 0.9 and 0.7 for offensive and hate speech tasks respectively.
\subsection{Multi-task learning based on Transfer learning:}
With the outstanding performances given by the transformers, most of the researchers have employed these pre-trained models in hate speech detection. 
Awal et al. \cite{awal2021angrybert} proposed “AngryBERT", a BERT-based multitask model that learns with sentiment analysis and target detection as auxiliary tasks. They demonstrated the ability of those latters to improve hate speech detection. Their model gave an F1 macro score of 90.71\% on the Davidson dataset \cite{davidson2017automated}. In addition, to address Aggression Identification, 
Samghabadi et al. \cite{samghabadi2020aggression} provided a neural model that builds attention on top of BERT using a multi-task learning paradigm. Their model scored 0.8579 weighted F1 on the English “Misogynistic Aggression Identification” task.
Moreover, using the pre-trained AraBERT, 
Djandji et al. \cite{djandji2020multi} enhanced this transformer with the inclusion of Multi-task learning to build a model able to learn well from little amounts of data, they trained it on several Arabic abusive speech datasets. Their model gave an F1 macro score of 90.15\% and 83.41\% in Offensive language and hate speech tasks respectively.
\subsection{Emotion knowledge to detect hate/offensive language}
In many fields, like the identification of mental diseases and social media analytics, understanding human emotional patterns is crucial. Since hate/offensive speech detection is integrally tied to the speaker's emotional state \cite{patrick1901psychology}, detecting emotions has also become an essential application. 
Markov et al. \cite{markov2021exploring} studied how stylometric and emotional characteristics affect the detection of hate speech. They demonstrated that, when integrated into an ensemble with deep learning models,
the use of those features surpasses the commonly used ones to detect hateful content.
 Moreover, Chiril et al. \cite{chiril2022emotionally} investigated the the affective knowledge extracted from Sentic computing resources and from structured hate lexicons. They implemented multitasking techniques, and they attained the greatest outcomes with models that used data from these affective resources.
Adding to that, Plaza-del-Arco et al. \cite{plaza2021multi} defended the hypothesis of the relationship between hate speech tasks and sentiments, emotions as well as targets, via a straightforward multi-task learning architecture. Implementing BERT-based multitask model, they got the best overall result of F1 = 0.79.
Furthermore, using a Transformer-based model, they also proposed a Multi-task model that makes use of shared sentiment and emotional knowledge to identify hate speech in Spanish tweets \cite{plaza2021multi1}. Their findings demonstrate that these knowledge work together to more reliably identify hate speech.
Based on the above-mentioned works, we propose several strategies for the same tasks by integrating the best of these approaches. Thus, we built a BERT-based and mBERT-based multi-task model using the knowledge extracted from emotion samples provided by social media platforms
using a large-size and more diverse-resources emotional dataset, and, implementing a shared representation to enable knowledge transfer between tasks and to reduce model complexity.


\section{Methodology}
\vspace{-5pt}
\subsection{Dataset}
We conducted our experiments on Davidson dataset (Twitter) \cite{davidson2017automated}: related to hateful and abusive language detection, and GoEmotion corpora (Reddit) \cite{demszky2020goemotions}: related to the emotion analysis. 
The corpora statistics are displayed in Table \ref{data}. 

\subsubsection{\textbf{Hate/Offensive speech dataset}}
For the training of our model for the hate/Offensive task, we used Davidson corpora \cite{davidson2017automated}. This data was compiled using a lexicon of hate speech content taken from Twitter and classified tweets into \textit{Hate speech}, \textit{Offensive}, and \textit{Neither} classes having about 24k total number of labeled samples. In this paper, we implement hate/offensive speech binary classification and thus, we created 2 corpora by separating the hateful labeled samples from offensive ones. As a result, we got Davidson-HATE and Davidson-OFF for hate speech and offensive language labeled datasets, respectively.
    
\subsubsection{\textbf{Emotion dataset}}
For training the emotion task, we use GoEmotions corpora \cite{demszky2020goemotions}, developed by Demzky et al. in 2020. This corpus is considered the largest manually annotated dataset, which consists of 58k English Reddit comments categorized as either Neutral or one of 27 emotion groups. The Ekman level further defines the emotion categorization into anger, disgust, fear, joy, sadness, and surprise. We implement the final corpora labeled as the Ekman model in our experiments in order to use a more generalizable dataset with less noisy data.


\begin{table}[htbp]
\begin{center}
\begin{tabular}{c c c}
\hline \myrowcolour%
\textbf{Dataset}                        & \textbf{Size}           & \textbf{Labels} \\ \hline \hline
\textbf{Davidson-HATE} & 5593  & hate \\
                                        &                         & normal          \\ \hline
\textbf{Davidson-OFF}  & 23 353 & offensive \\
                                        &                         & normal          \\ \hline
\textbf{GoEmotions}    & 48 834 & anger, disgust, fear, \\
& & joy, surprise, sadness, neutral\\            
\end{tabular}
\end{center}
\caption{General overview of the datasets along with their number of samples and labels' distributions.}
\vspace{-18pt}
\label{data}
\end{table}

\begin{figure*}[t]
\centering
\includegraphics[width=7in]{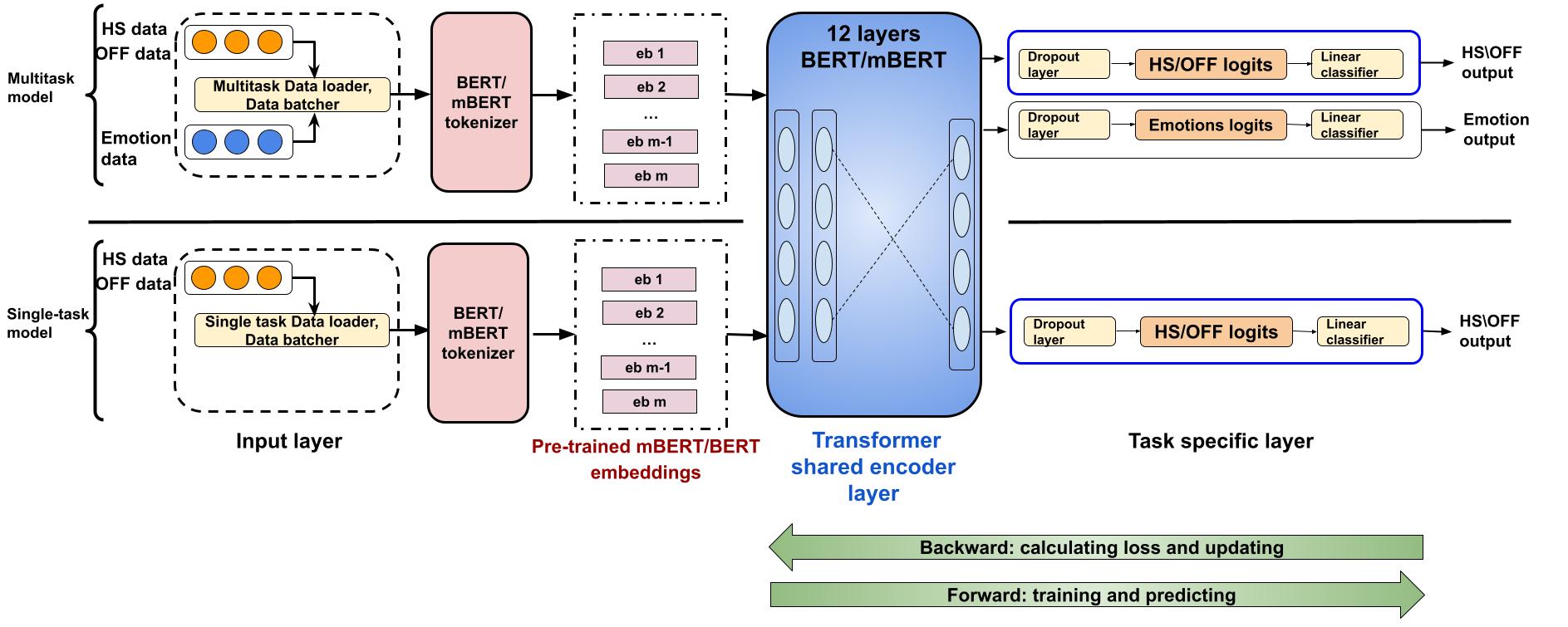}
 \caption{Overview of the BERT and mBERT-based:  Single-task, and Multitask leveraging emotions representations (auxiliary task) model architecture in the detection of Hate speech  “HS" and Offensive language  “OFF" (main tasks) from text input}
\label{mtl}
\end{figure*}
\subsection{Transformer-based multi-task approach MTL}
Unlike Single Task Learning (STL) which learns task-specific features from one dataset at a time, Multi-Task Learning (MTL) aims to tackle multiple issues at once. 
In STL, input vectors $x_{i}$ are supervisedly mapped to any label $y_{i}$ in order to train the model to complete a classification task $T$. Each sentence $x_{i}$ is processed through the model layers, and the final representation is then run through Softmax to predict the probability distribution over $C$ classes. Given a dataset $D$ with $n$ training labeled inputs $(xi, yi)$, the models's weights are  
trained in order to reduce the cross-entropy of the detected $y^{\hat{}}$ and labeled $y$ samples.
Where, 
\vspace{-5pt}
\begin{align}
    y^{\hat{}} = softmax(W + b)
\end{align}

Knowing that, $W$ is the final
weight after optimization of the linear classifier when training, and $b$ is a bias term \cite{kapil2020deep}.
The objective of MTL scenario is to employ the method of learning numerous tasks to enhance performance on each one of them \cite{caruana1997multitask}. Although they could have different data or characteristics, these tasks are correlated and share some similarities. And when the model is trained, it can exploit shared characteristics by using some hints from one task to enhance the other ones. 
To understand more the process of building an MTL model, Zhang et al. \cite{zhang2021survey} defined MTL as follows:
Given $n$ learning tasks,
$\left\{ T_{i}\right\}_{i=1}^{n}$,
MTL seeks to enhance the learning of a model for the classification task $T_{i}$ by leveraging the knowledge in some or all of the $n$ tasks, where all or a subset of the tasks are connected.
The two most popular methods to share knowledge in multi-tasking are  hard parameter sharing and soft parameter sharing\cite{xue2007multi}.
\textbf{Hard parameter sharing} involves all tasks sharing the hidden layers with a number of task-specific output layers, which is the method we used by implementing shared BERT and mBERT encoders, 
in order to avoid task-specific parameters for each task, because it can lead to augmented model complexity \cite{liu2019auxiliary}.
On the other side, each task can have its own layers with certain shareable components, known as \textbf{soft parameters sharing}.
Overall MTL approach efficiently improves the sample size when training a model, which leads to enhanced performance by improving the generalization of model in comparison to learning several tasks separately \cite{zhang2021survey}.

In this study, we used three related tasks: Hate speech detection, Offensive language detection, and Emotion recognition. The purpose is to determine whether implementing an MTL scenario to emotion categorization task facilitates the identification of hate/offensive speech, regardless of the source of social media data. 
Therefore, we build a typical contextualized embedding configuration where  the input is represented by a well-known model, the Bidirectional Encoder Representations from Transformers (BERT) \cite{devlin2018bert} and the Multilingual BERT (mBERT) \footnote{\url{https://huggingface.co/bert-base-multilingual-uncased}\label{mbert}}.
We used the latter model in order to build a cross-lingual generalizable approach, it could be tested on different target languages or used to be trained on monolingual low-resource ones. We used it also to compare with BERT based MTL model to understand more about the influence of adding features from other languages. 
We added two sequence classification heads to the encoder, one for Hate/offensive speech and another for Emotion recognition. 
The two tasks jointly share the transformer encoder, as seen in Figure \ref{mtl}
, so that one task profits from the other by sharing features. The importance of using the shared encoder is to guarantee that any adjustments on its weights during training will change the same encoder weights, and not to use any extra GPU memory.
The output heads for every task are then generated, and each task head is connected to a common sentence encoder.
The layers are then adjusted in accordance with the specified set of our downstream tasks.
As shown in Figure \ref{mtl}, the input representation is BERT/mBERT-based tokenization, and each task corresponds to a specific classification head. In the first step, given a data input, it is first tokenized using the default tokenizer of BERT/mBERT
and then converted into pre-trained BERT embeddings: $Eb = \left \{ eb_{1} , eb_{2}, ..., eb_{n} \right \}$.
These embeddings are then sent to the pre-trained BERT/mBERT shared encoder. 
After defining the feature extraction function using the corresponding tokenizer, we utilize the \textit{“dataset.map"} method from the NLP package to apply this function to our data inputs. This NLP library effectively manages the mapping and caches the features.
We constructed a \textit{ “MultitaskDataloader"} that combines many data loaders (built to load data of each task) into a single loader. The latter aims to sample, randomly, data from these data loaders, build a task batch and produce the associated task name (attached to each batch data). 
Overall, information can flow from one task head to another through the shared encoder, which, is updated during the training via backpropagation. In fact, BERT/mBERT gets tuned by the combined loss of both tasks (Cross-Entropy loss) in order to learn a shared set of information between both tasks.
As for the task-specific layers these consist of a task-specific softmax activation followed by a linear classification layer that is dedicated to extracting the unique information per task and giving final outputs.

\section{Experiments and Results}
\vspace{-5pt}
\subsection{Data preprocessing}
We use the below procedures to pre-process the Twitter dataset using the Ekphrasis library \cite{baziotis-pelekis-doulkeridis:2017:SemEval2}:
1) Switching into lowercase, 2) Delete URLs and emails, 3) Remove users' names and mentions, 4) Shorten prolonged words and delete repeated characters (“yeeessss" to “yes"...), 5) Keep stop words, 6) Remove punctuations, unknown uni-codes, and additional delimiting characters, 7) Remove hashtags (\#) and correct their texts (e.g, “\#notracism" to “not racism"), 8) Eliminate tweets of length less than 2, and 9) Remove emojis.

\begin{table*}[]
\centering
\vspace{0.7in}
\resizebox{18cm}{!}{\begin{tabular}{|cl|ccccc|ccccc|}
\hline \myrowcolour%
\multicolumn{2}{|c}{}                                                                                                                               & \multicolumn{5}{|c|}{\textbf{Hate speech detection HS}}                                                                                                                        & \multicolumn{5}{c|}{\textbf{Offensive language detection OFF}}                                                                                                                \\ \cline{3-12} \myrowcolour%
\multicolumn{2}{|c}{\textbf{Model}}                                                                                                                                                & \multicolumn{1}{|c}{\textbf{Acc.}} & \multicolumn{1}{c}{\textbf{Pr.}} & \multicolumn{1}{c}{\textbf{Recall}} & \multicolumn{1}{c}{\textbf{F1(m)}} & \textbf{F1(w)} & \multicolumn{1}{c}{\textbf{Acc.}} & \multicolumn{1}{c}{\textbf{Pr.}} & \multicolumn{1}{c}{\textbf{Recall}} & \multicolumn{1}{c}{\textbf{F1(m)}} & \textbf{F1(w)} \\ \hline \hline
\multicolumn{1}{|c|}{\textbf{Baselines}} & \textbf{\begin{tabular}[c]{@{}l@{}} BERT-based ensemble \cite{mnassri2022bert}\end{tabular}} & \multicolumn{1}{c}{\textbf{0.9474}}        & \multicolumn{1}{c}{\textbf{0.9346}}       & \multicolumn{1}{c}{\textbf{0.9234}}          & \multicolumn{1}{c}{\textbf{0.9288}}            & \multicolumn{1}{c}{\textbf{0.9470}}              & \multicolumn{1}{|c}{-}              & \multicolumn{1}{c}{-}             & \multicolumn{1}{c}{-}                & \multicolumn{1}{c}{-}                     & \multicolumn{1}{c|}{-}                  \\ \cline{2-12} 
\multicolumn{1}{|c|}{}                                    & \textbf{BERT-MLP \cite{mnassri2022bert}}                                                                                         & \multicolumn{1}{c}{-}              & \multicolumn{1}{c}{-}             & \multicolumn{1}{c}{-}                & \multicolumn{1}{c}{-}            & \multicolumn{1}{c|}{-}                            & \multicolumn{1}{c}{0.9682}        & \multicolumn{1}{c}{0.9400}       & \multicolumn{1}{c}{0.9532}          & \multicolumn{1}{c}{0.9465}            & \multicolumn{1}{c|}{0.9685}               \\ \cline{2-12} 
\multicolumn{1}{|c|}{}                                    & \textbf{\begin{tabular}[c]{@{}l@{}}BERT STL \end{tabular}}                        & \multicolumn{1}{c}{0.9165}        & \multicolumn{1}{c}{0.8956}       & \multicolumn{1}{c}{0.8788}          & \multicolumn{1}{c}{0.8867}            & 0.9156               & \multicolumn{1}{c}{0.9689}        & \multicolumn{1}{c}{0.9400}       & \multicolumn{1}{c}{0.9566}          & \multicolumn{1}{c}{0.9480}            & 0.9692                                                     \\ \cline{2-12} 
\multicolumn{1}{|c|}{}                                    & \textbf{\begin{tabular}[c]{@{}l@{}}mBERT STL \end{tabular}}                      & \multicolumn{1}{c}{0.9275}        & \multicolumn{1}{c}{0.9109}       & \multicolumn{1}{c}{0.8934}          & \multicolumn{1}{c}{0.9016}            & 0.9268              & \multicolumn{1}{c}{0.9645}        & \multicolumn{1}{c}{0.931}        & \multicolumn{1}{c}{0.952}           & \multicolumn{1}{c}{0.9410}            & 0.9649               \\                                \hline \hline

\multicolumn{2}{|c|}{\textbf{BERT MTL}}                                                                                                       & \multicolumn{1}{c}{0.9385}        & \multicolumn{1}{c}{0.9365}       & \multicolumn{1}{c}{0.8971}          & \multicolumn{1}{c}{0.9146}            & 0.9371

& \multicolumn{1}{c}{\textbf{0.9691}}        & \multicolumn{1}{c}{\textbf{0.9367}}       & \multicolumn{1}{c}{\textbf{0.9625}}          & \multicolumn{1}{c}{\textbf{0.9489}}            & \textbf{0.9695 }              \\ \hline

\multicolumn{2}{|c|}{\textbf{mBERT MTL}}                                                                                                       & \multicolumn{1}{c}{\textbf{0.9413}}        & \multicolumn{1}{c}{\textbf{0.9294}}       & \multicolumn{1}{c}{\textbf{0.9123}}          & \multicolumn{1}{c}{\textbf{0.9204}}            &\textbf{ 0.9407 }            & \multicolumn{1}{c}{0.9674}        & \multicolumn{1}{c}{0.9346}       & \multicolumn{1}{c}{0.9558}          & \multicolumn{1}{c}{0.9459}            & 0.9678               \\ \hline
\end{tabular}}
\vspace{1pt}
\caption{Single-Task Learning - STL vs Multi-Task Learning - MTL models for Hate speech (HS) and Offensive language (OFF) detection. \\
\textit{BERT-based ensemble: Soft average voting Ensemble+BERT-CNN+BERT-LSTM, and MLP - Multi Layer Perceptron. Acc. - Accuracy and Pr. - Precision, F1(m) - F1 Macro and F1(w) - F1 Weighted.}}
\label{results}
\end{table*}
\subsection{Data analysis platform and evaluation metrics}
The MTL models have been implemented using PyTorch. We trained the models on the training set and tested them on the validation set, keeping the same data split for GoEmotion corpora, and partitioning Davidson into 80\% train and 20\% validation set.
The implemented models are trained using batch size 8 on Google Colab Pro (Tesla-T4 GPU environment with 32 GB of RAM).
We used an optimizer with a learning rate of 1e-5, and experimented with the Cross-entropy loss function.
Since we used imbalanced datasets,
classifiers' performances are measured via multiple metrics: macro and weighted averaged F1 scores, precision, recall as well as accuracy.
Weighted F1 calculates the score for each class and adds them together using a \textbf{weight} that depends on the number of true labels of each class. Because we were using an imbalanced dataset, we want to assign more contributions to classes having more samples.
\subsection{Results and interpretations}
In this section, we compare the performance of single-task models with multi-task ones to understand the importance of using external, but related, features (emotions) in the abusive language detection.
The STL models proposed in our previous contribution achieved considerable performances for detecting abusive content \cite{mnassri2022bert} where the ensemble average voting of BERT-CNN+BERT-LSTM gave better results in HS task and BERT-MLP (BERT-Multi Layer Perceptron) for the OFF task. Hence, we use these two best-performed models in \cite{mnassri2022bert}, as well as BERT and mBERT as our single-task baseline models to compare with MTL approaches. 
Furthermore, we carried out an error analysis to get more information regarding the performance of the suggested MTL models. Working on imbalanced datasets, we want to get deeper into the classification of each class. So, we analyzed the confusion matrix and compared the MTL misclassification error with the other models as illustrated in the table \ref{results}.
Even though we have an emotion classifier as shown in Figure \ref{mtl}, this work mainly focuses on hate speech and offensive language detection tasks. Hence, Table \ref{results} illustrates our experimental results for hate speech detection task \textit{“HS task"} and offensive language detection task \textit{“OFF task"}. In fact, the emotional analysis task trains the MTL network how to identify the emotion labels from the input samples and the representations generated by the encoder including the affective knowledge. This enables the MTL model to detect HS and OFF more accurately by leveraging the affective nature of the text input. 
Overall, the STL and MTL models' results in both tasks (hate speech detection and offensive language detection) reveal a good performance when fine-tuning on small, imbalanced datasets. 
\subsubsection{Hate speech detection}
\begin{figure}
    \centering
    \subfigure a. {\includegraphics[width=0.2\textwidth]{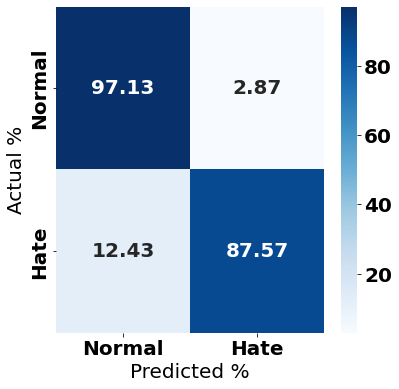}}\label{a}
    \hspace{-1\baselineskip}
    \subfigure b.{\includegraphics[width=0.2\textwidth]{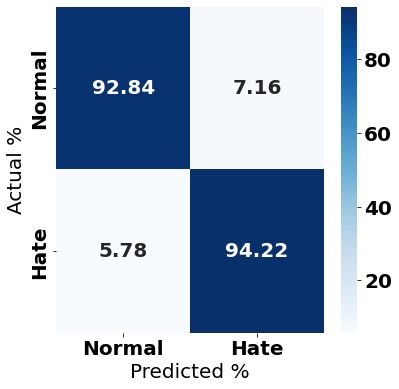}} \label{b}
 \vspace{-1pt}
    \caption{Confusion matrix of Hate speech detection task:\\ 
    (a.) BERT-based ensemble model. (b.) mBERT muti-task model.}\label{hscm}
\end{figure}

As illustrated in Table \ref{results}, for HS detection task, the performance of the mBERT-MTL model succeeded in surpassing all the STL models except the BERT-based ensemble model. Compared with the BERT STL model, these two models increased their performances by 3\% and interestingly, there is no significant difference in their performances. For instance, the accuracy of the HS task for the ensemble model and mBERT-MLT model are 0.9474 and 0.9413, respectively and that of F1 macro are 0.9288 and 0.9204, respectively. 
Therefore, in order to understand the best model out of the BERT-based ensemble model vs mBERT MTL model in HS task, we conducted an error analysis based on the confusion matrix to explore which model exhibits less misclassification errors. Thus, we compare the confusion matrix of mBERT-MTL and BERT-based ensemble model as displayed in Figure \ref{hscm}. For HS task, although the ensemble model outperforms mBERT-MTL, we noted that the latter can detect Hate speech samples more efficiently than the Ensemble model, which is worth noticing since we used an imbalanced Hate speech dataset with only $\sim$77\%
of samples labeled as “hate". This indicates that the MTL model has less misclassification errors compared to the ensemble one, getting fewer percentages of false positives and false negatives for both classes.

\subsubsection{Offensive language detection}

\begin{figure}
    \centering
    \subfigure a.{\includegraphics[width=0.2\textwidth]{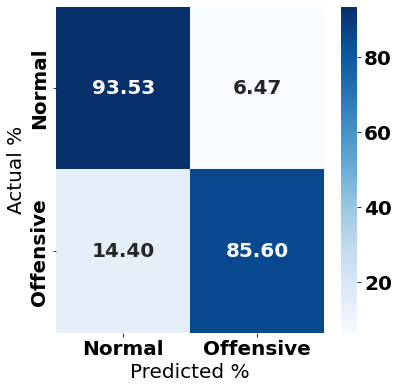}}\label{a}
    \hspace{-1\baselineskip}
    \subfigure b.{\includegraphics[width=0.2\textwidth]{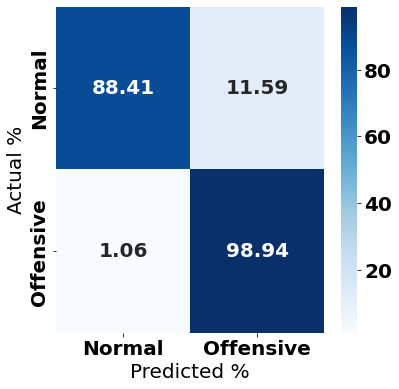}} \label{b}
   \vspace{-5pt}
    \caption{Confusion matrix of Offensive language detection task: \\
    (a.) BERT single-task model, (b.) BERT multi-task model.}\label{offcm}
\end{figure}

We can interpret from Table \ref{results} that, in OFF task, BERT-MTL gave the highest performances. However, we observe that, compared to the other mBERT-MTL model and the single-task models, the performance improvement is not very significant. The main reason behind this is due to the high imbalance Davidson-OFF corpora ($\sim$91\% offensive samples) \cite{chen2021multi}. Nevertheless, emotional features help to improve the OFF task classification by a small margin indicating that even with the highly imbalanced dataset, external features increase model performance.  
Further, we observed misclassification errors using the confusion matrix of BERT-MTL and BERT-STL models, the two best-performed models in the OFF task, as shown in Figure \ref{offcm}. 
BERT-MTL illustrates its ability to correctly detect offensive content with the largest rate among all baseline models, leading to True positives of 98.94\% for “offensive" class, as well as the least misclassification error rate. In addition, the false positive rate of BERT-MTL model (1.06\%) is fewer compared to BERT-STL model(14.40\%). This indicates that multi-task models contributed to correctly classifying the texts in comparison to single-task models. Similar to the HS task, emotional features help to improve OFF task performance. 
Overall, based on the results obtained by the proposed MTL models for both hate speech and offensive language detection tasks, multi-task joint learning models outperformed single-task models and exhibit less false positive errors, indicating that emotional knowledge helps to improve the classification. 



\vspace{-5pt}


\section{Conclusion and Future Work}
\vspace{-5pt}
Recent years have seen a rise in the dissemination of abusive language, making it a significant issue for state governments and social media corporations to find and delete this kind of content.
As a result, we focus our paper on Hate speech and offensive language detection through a BERT/mBERT-based multi-task model to benefit from emotion analysis as a related task.
Due to the sensitivity and granularity of hate speech and offensive language, we conducted experiments on two datasets extracted from Davidson corpora to consider, separately, hate speech and offensive language detection, each one, as a major classification task.
The efficiency of our suggested model (in terms of performance and resource consumption) and a thorough investigation of the transfer of affective knowledge, demonstrate how emotion classification tasks enable the multi-task system to predict hate/offensive language more precisely by leveraging on this associated information.
To improve the multi-task approach, we can focus more on the training datasets used, mainly by reducing the imbalance ratio through different data augmentation techniques to over-sample the corpora.
As for the related external features, hate/offensive language detection task is not restricted only to emotion analysis, but can also be integrated with polarity, target (towards individuals or groups), irony, or sarcasm detection tasks. Thus we consider further using other related features. 
Furthermore, by implementing mBERT, and getting good performance, we can measure the cross-lingual generalization of our approach using zero-shot or few-shot learning and test it on several target low-resource languages (Arabic, French, German, etc.). We also aim to propose feature fusion approaches for hate speech detection as a joint learning approach using fuzzy ruling. In addition, we aim to compare different model performances in terms of resource consumption (i.e., memory, run-time) to determine the most optimized solution to deploy in real environments. 

\end{document}